%% file: sample-sigconf.tex
\documentclass[sigconf]{acmart}
\AtBeginDocument{%
  }

\setcopyright{acmlicensed}
\copyrightyear{2018}
\acmYear{2018}
\acmDOI{XXXXXXX.XXXXXXX}
\acmConference[Conference acronym 'XX]{Make sure to enter the correct
  conference title from your rights confirmation email}{June 03--05,
  2018}{Woodstock, NY}
\acmISBN{978-1-4503-XXXX-X/2018/06}

\usepackage{booktabs}
\usepackage{bm}
\usepackage{multirow}
\usepackage{algorithm}
\usepackage{algpseudocode}
\usepackage{amsmath}
\usepackage{listings}

\definecolor{c1}{RGB}{192, 0, 0}
\definecolor{c2}{RGB}{0, 112, 192}
\definecolor{c3}{RGB}{0, 0, 0}

\newcommand{\first}[1]{\textbf{{\color{c1}#1}}}
\newcommand{\second}[1]{{\textbf{\color{c2}#1}}}

\begin{document}

%%
%% The "title" command has an optional parameter,
%% allowing the author to define a "short title" to be used in page headers.
\title{Scalable and Efficient Joint Spiking Embedding Predictive Architecture for Large-Scale Dynamic Graphs}

%%
%% The "author" command and its associated commands are used to define
%% the authors and their affiliations.
%% Of note is the shared affiliation of the first two authors, and the
%% "authornote" and "authornotemark" commands
%% used to denote shared contribution to the research.
% \author{Anonymous Authors}

%%
%% By default, the full list of authors will be used in the page
%% headers. Often, this list is too long, and will overlap
%% other information printed in the page headers. This command allows
%% the author to define a more concise list
%% of authors' names for this purpose.
% \renewcommand{\shortauthors}{Trovato et al.}
\author{Huizhe Zhang}
\affiliation{%
  \institution{Sun Yat-sen University}
  \city{Guangzhou}
  \country{China}
}
\email{zhanghzh33@mail2.sysu.edu.cn}

\author{Yuchang Zhu}
\affiliation{%
  \institution{Sun Yat-sen University}
  \city{Guangzhou}
  \country{China}}
\email{zhuych27@mail2.sysu.edu.cn}

\author{Huazhen Zhong}
\affiliation{%
  \institution{Sun Yat-sen University}
  \city{Guangzhou}
  \country{China}}
\email{zhonghzh9@mail2.sysu.edu.cn}

\author{Liang Chen}
\authornote{Corresponding author.}
\affiliation{%
  \institution{Sun Yat-sen University}
  \city{Guangzhou}
  \country{China}}
\email{chenliang6@mail.sysu.edu.cn}

\author{Zibin Zheng}
\affiliation{%
  \institution{Sun Yat-sen University}
  \city{Zhuhai}
  \country{China}}
\email{zhzibin@mail.sysu.edu.cn}

%%
%% The abstract is a short summary of the work to be presented in the
%% article.
\begin{abstract}
Dynamic graph learning aims to capture evolving structural and semantic patterns in real-world systems, such as fraud detection and recommender systems. Due to the scarcity of labeled data in real-world dynamic graphs, recent studies have introduced generative or contrastive paradigms (e.g., masked graph autoencoders or graph contrastive learning) to generate task-agnostic graph embeddings. However, these methods typically rely on complex edge-level reconstruction objectives and tailored graph augmentation strategies. It incurs substantial computational overhead when scaling to large-scale dynamic graphs. 
In this paper, we propose \textbf{SG-JEPA}, a joint spiking embedding predictive architecture for large-scale dynamic graphs. In contrast to existing self-supervised methods, SG-JEPA partitions nodes into context and target sets along the temporal dimension to learn embeddings that are predictive of each other via additional spatial-temporal information. Furthermore, through encoding sequential inputs into coarse-to-fine spike count embeddings, spiking neurons enable SG-JEPA to adapt to the varying computational constraints of downstream tasks. Extensive experiments demonstrate that SG-JEPA achieves competitive or even superior performance over discriminative baselines on node classification, while effectively scaling to the dynamic graph with 13 million edges. SG-JEPA avoids the complex machinery (negative sampling, graph augmentations, edge-level reconstruction, etc.), resulting in superior training efficiency and memory scalability compared with prior self-supervised dynamic graph baselines.
\end{abstract}

%%
%% The code below is generated by the tool at http://dl.acm.org/ccs.cfm.
%% Please copy and paste the code instead of the example below.
%%


%%
%% Keywords. The author(s) should pick words that accurately describe
%% the work being presented. Separate the keywords with commas.
%% A "teaser" image appears between the author and affiliation
%% information and the body of the document, and typically spans the
%% page.
% \begin{teaserfigure}
%   \includegraphics[width=\textwidth]{sampleteaser}
%   \caption{Seattle Mariners at Spring Training, 2010.}
%   \Description{Enjoying the baseball game from the third-base
%   seats. Ichiro Suzuki preparing to bat.}
%   \label{fig:teaser}
% \end{teaserfigure}

\received{20 February 2007}
\received[revised]{12 March 2009}
\received[accepted]{5 June 2009}

%%
%% This command processes the author and affiliation and title
%% information and builds the first part of the formatted document.
\maketitle

\input{sec/01_intro}
\input{sec/02_relate}
\input{sec/03_method}
\input{sec/04_exp}
\input{sec/05_concl}

\bibliographystyle{ACM-Reference-Format}
\bibliography{sample-base}

\end{document}

%% file: sec/01_intro.tex
\section{Introduction}
Graphs are a fundamental data structure for modeling complex relational systems, ranging from social interactions to traffic systems. In many real-world scenarios, some systems evolving over time can be modeled as dynamic graphs that encode both structural and temporal dependencies. To capture such evolving patterns, Dynamic Graph Neural Networks (DGNNs) are proposed to integrate graph representation learning with temporal modeling, which have achieved strong performance across tasks such as link prediction and node classification \cite{skarding2021foundations, feng2025comprehensive}. Different from vanilla static GNNs, DGNNs continuously update node representations by incorporating temporal information from evolving interactions or graph snapshots \cite{yu2023towards, li2024state, zhu2024topology}. However, most DGNNs tailored for supervised or semi-supervised settings rely heavily on labeled data, which is often scarce or expensive to obtain in real-world applications. 

\begin{figure*}[!ht]
  \begin{center}
    \centerline{\includegraphics[width=0.9\linewidth]{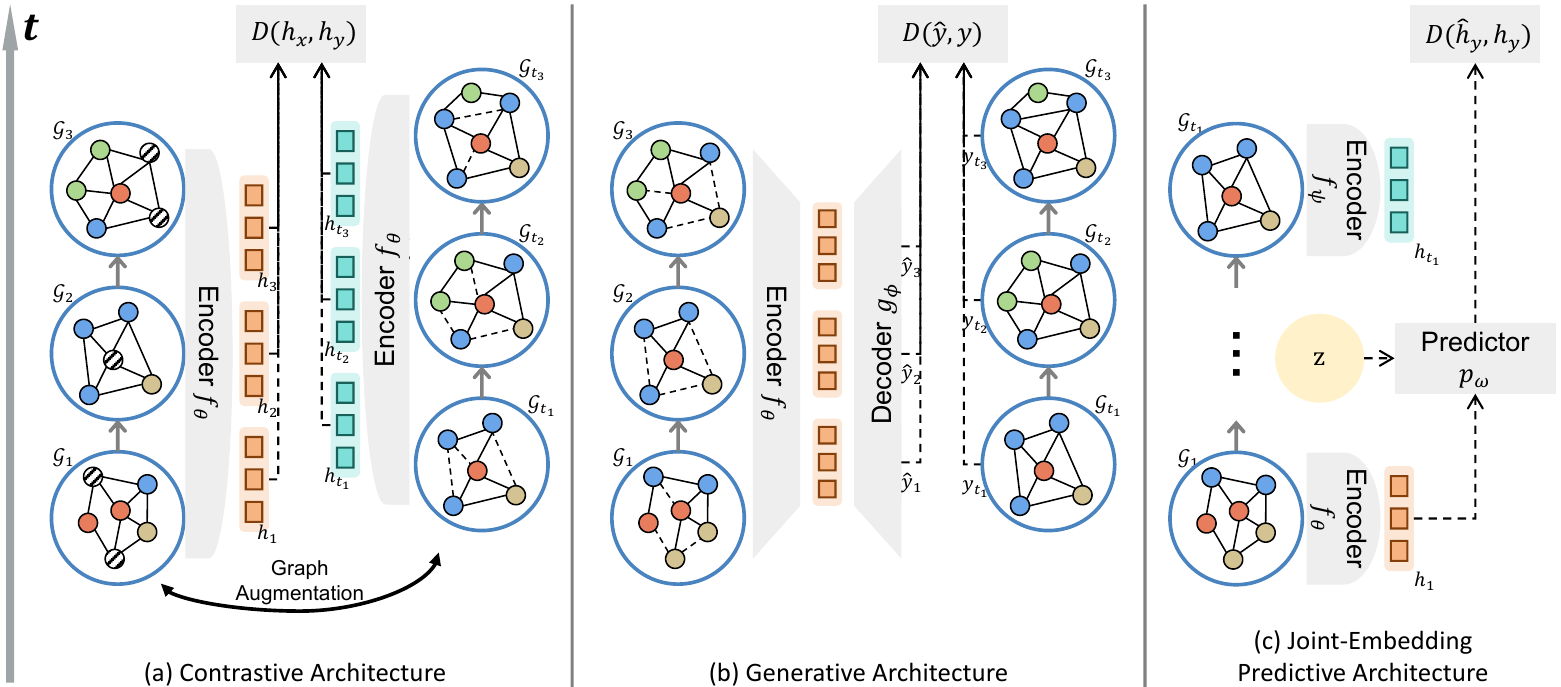}}
    \caption{Illustration of self-supervised learning frameworks for DGNNs. (a) Contrastive-based methods construct multiple augmented temporal graph views and align representations across different views via contrastive objectives, which rely on well-designed graph augmentation strategies. (b) Generative-based methods learn representations by reconstructing masked edges or temporal signals through decoder networks. They often incur high computational and memory overhead on large-scale dynamic graphs. (c) Joint-embedding predictive architecture models temporal evolution by predicting target node representations from historical context representations directly in embedding space.}
    \label{fig:ssl}
  \end{center}
\end{figure*}

This limitation motivates the exploration of self-supervised paradigms for dynamic graph representation learning. Recent efforts attempt to alleviate label dependency by introducing contrastive or generative objectives for dynamic graphs. Contrastive methods construct multiple graph views through structural perturbations or temporal subgraph sampling, and then maximize agreement between representations of positive pairs while repelling negatives \cite{jiang2021self, alomrani2022dyg2vec, chen2023self}. Although effective, such approaches typically rely on carefully designed graph augmentations and negative sampling strategies. As shown in Figure~\ref{fig:ssl}, it introduces substantial computational overhead and inductive biases. Inspired by masked autoencoders, generative methods tend to reconstruct masked edges or temporal signals from partial observations \cite{he2025maskdgnn}. While they avoid explicit negative sampling, reconstruction-based objectives often emphasize low-level structural fidelity and incur computational cost when scaling to large dynamic graphs. More importantly, the choice of effective augmentations (e.g., edge modification or node feature masking) is often task-dependent and requires careful manual selection, which limits adaptability and scalability in real-world dynamic graphs.

Joint-Embedding Predictive Architecture (JEPA), as an emerging alternative, was recently proposed in computer vision as a non-generative framework for learning semantic representations \cite{assran2023self, bardes2023v, assran2025v}. Instead of reconstructing raw inputs or enforcing invariance across augmented views, JEPA learns representations that are predictive of each other when conditioned on additional latent information. By performing prediction directly in representation space, JEPA avoids pixel-level reconstruction and reduces reliance on handcrafted augmentations, leading to improved scalability and semantic abstraction. 
Although JEPA has demonstrated strong performance in image modeling, its potential in dynamic graph learning remains largely underexplored. We believe that the predictive learning principle of JEPA naturally aligns with the temporal evolution in dynamic graphs, where representations at different time steps should be mutually predictive under spatio-temporal conditioning. However, directly extending it to dynamic graph representation learning is non-trivial.We identify two key challenges in applying JEPA to dynamic graphs. 
First, it remains unclear how to construct effective context–target partitions on evolving graph structures, where node neighborhoods and temporal dependencies continuously change. 
Second, JEPA critically depends on an appropriate encoder to model spatio-temporal dynamics on dynamic graphs. Most existing dynamic graph neural networks are constrained by limited scalability. For example, recurrent designs require maintaining dense hidden states for each node over time, which leads to substantial memory and computational overhead, making them impractical for large-scale dynamic graphs.

To address these challenges, we propose \textbf{SG-JEPA}, a Spiking Joint-Embedding Predictive Architecture for large-scale dynamic graph representation learning. Instead of reconstructing edges or enforcing invariance across augmented views, SG-JEPA formulates dynamic graphs as a temporal prediction problem by learning to predict future node representations from historical context representations in embedding space. Specifically, SG-JEPA partitions nodes along the temporal dimension into context and target sets, and learns representations that are mutually predictive under spatio-temporal conditioning. SG-JEPA further employs a time-aware spiking encoder to efficiently capture spatio-temporal evolution. This design enables scalable predictive learning on large dynamic graphs while avoiding the computational bottlenecks of conventional dynamic representation learning methods.
Our contributions are summarized as follows:
\begin{itemize}
    \item We are the first to introduce JEPA into dynamic graph representation learning. We propose a temporal context–target formulation that predicts future node representations from historical context representations in embedding space.
    \item We develop a spiking time-aware predictive module that encodes dynamic graph events into coarse-to-fine spike-count embeddings, enabling adaptive and resource-efficient computation for downstream tasks.
    \item Extensive experiments demonstrate that SG-JEPA achieves competitive or superior performance compared to supervised or semi-supervised baselines on node classification tasks, while significantly improving training efficiency.
\end{itemize}

%% file: sec/02_relate.tex
\section{Related Work}
\subsection{Dynamic GNNs}
Dynamic Graph Neural Networks (DGNNs) extend static graph representation learning to evolving relational data by modeling both structural dependencies and temporal dynamics. Existing dynamic graph models can be broadly categorized into Discrete-Time Dynamic Graphs (DTDG) and Continuous-Time Dynamic Graphs (CTDG). Discrete-time methods represent dynamic graphs as sequences of snapshots and apply recurrent or attention-based modules to capture temporal evolution across snapshots \cite{pareja2020evolvegcn, zhu2023wingnn, chen2025signn}. For example, VGRNN \cite{hajiramezanali2019variational} extends variational graph autoencoders to dynamic graphs by integrating recurrent neural networks to model temporal dependencies, learning evolving node representations through probabilistic latent variables. Continuous-time methods instead treat interactions as time-stamped events and update node representations through event-driven mechanisms or memory modules \cite{yu2023towards, ma2020streaming, kumar2019predicting}. Wang et al. \cite{wang2021inductive} propose Causal Anonymous Walks to inductively represent temporal networks by extracting anonymized temporal random walks, which enables efficient and generalizable link prediction. 
Recent works have explored self-supervised paradigms for dynamic graphs. Gao et al. \cite{gao2025dvgmae} extend masked autoencoders with a temporally aware edge-masking strategy and a globally enhanced decoder to jointly capture evolving behaviors and topological structures in dynamic graphs. DDGCL \cite{tian2021self} proposes the first self-supervised contrastive learning framework for dynamic graphs through a time-dependent similarity function and a debiased GAN-style contrastive loss. 
Existing self-supervised dynamic graph methods mainly rely on contrastive or generative objectives with graph augmentations or reconstruction. In contrast, we explore a predictive learning paradigm that models temporal dependencies via latent-space representation prediction.

\subsection{Spiking Neural Networks} Spiking Neural Networks (SNNs) have been widely studied as a biologically inspired and energy-efficient alternative to conventional artificial neural networks. SNNs operate via discrete spike events, where information is encoded in spike timing and firing patterns \cite{eshraghian2023training}.
Building upon SNNs, a line of recent studies has explored spiking graph neural networks (SGNNs) to extend spiking computation to graph-structured data. Extensive experiments demonstrate their effectiveness in capturing temporal dependencies and improving scalability on dynamic graphs \cite{zhu2022spiking, sun2024spiking, zhang2025gt}. Yao et al. \cite{yao2023attention} introduce a multi-dimensional attention mechanism for SNNs that modulates membrane potentials across temporal channels and spatial dimensions, which effectively narrows the performance gap between SNNs and ANNs. Dy-SIGN \cite{yin2024dynamic} compensates information loss via cross-layer propagation and reduces memory overhead by extending implicit differentiation to learn the dynamic graph representations. 
However, existing works largely integrate SNNs within supervised or contrastive frameworks. The synergy between spiking temporal dynamics and predictive representation learning in dynamic graphs remains underexplored, motivating our integration of SNNs within a joint-embedding predictive paradigm.

\subsection{Joint Embedding Predictive Architecture} Joint-Embedding Predictive Architectures (JEPAs) are a class of non-generative self-supervised learning methods that operate directly in embedding space by predicting the representation of a target view from that of a context view rather than reconstructing raw inputs or contrasting augmented pairs \cite{assran2023self}.
JEPA-style frameworks have shown improved scalability and semantic abstraction compared to generative or contrastive objectives \cite{assran2025v, li2025rethinking, xu2025next}. For example, VL-JEPA \cite{chen2025vl} is a vision–language model that replaces autoregressive token generation with JEPA-style latent embedding prediction. It achieves more efficient and versatile vision–language understanding and generation with fewer parameters and adaptive decoding. Fei et al. \cite{fei2023jepa} extend the JEPA to audio by predicting masked spectrogram region embeddings in latent space with time–frequency-aware curriculum masking. C-JEPA \cite{abdelfattah2024s} enhances I-JEPA by integrating VICReg-style variance–invariance–covariance regularization to prevent representation collapse and improve mean prediction, resulting in more stable and higher-quality self-supervised visual representations.

In this work, we extend the JEPA to dynamic graph representation learning by constructing temporal context–target partitions on evolving graphs and incorporating spiking-based temporal modeling, enabling efficient predictive learning under dynamic and large-scale settings.

\section{Preliminaries}
\paragraph{Dynamic Graph.}
We model a dynamic graph as a sequence of discrete-time graph snapshots
$\mathcal{G} = \{\mathcal{G}_1, \mathcal{G}_2, \ldots, \mathcal{G}_T\}$,
where each snapshot $\mathcal{G}_t = (\mathcal{V}_t, \mathcal{E}_t)$
represents the graph structure observed at time step $t$.
Here, $\mathcal{V}_t \subseteq \mathcal{N}$ denotes the set of nodes active at time $t$. Each node $u \in \mathcal{V}_t$ is associated with a node feature $\mathbf{x}_{u,t} \in \mathbb{R}^{d_x}$. $\mathcal{E}_t \subseteq \mathcal{V}_t \times \mathcal{V}_t$ denotes the set of edges at that snapshot.

\paragraph{Spiking Neural Network.} 
To enable efficient temporal modeling with biologically inspired dynamics, we adopt the Parametric Leaky Integrate-and-Fire neuron (PLIF) as the basic computational unit of SNNs \cite{fang2021incorporating}, which is characterized by three core operations: \textbf{Integrate} \textbf{Fire} and \textbf{Reset}. 
At each discrete time step $t$, the membrane potential $V^t$ integrates the input current $I^t$, emits a spike when exceeding a threshold, and is reset accordingly:
\begin{align}
    V^t &= V^{t-1} + \frac{1}{1+\exp(-\beta)}(I^t-(V^{t-1}-V_{reset})), \\
    S^t &= \Theta(V^t - V_{\mathrm{th}}), \\
    V^t &= V^t (1 - S^t) + V_{\mathrm{reset}} S^t ,
\end{align}
where $S^t \in \{0,1\}$ denotes the emitted spike, $V_{\mathrm{th}}$ is the firing threshold, and $V_{\mathrm{reset}}$ is the reset potential.
For notational simplicity, we denote the above membrane potential update and spike generation process by a unified operator $\mathrm{SNN}(\cdot)$, hereafter. Since the spike function $\Theta(\cdot)$ is non-differentiable, we adopt surrogate gradient methods to enable gradient-based optimization.
Specifically, during backpropagation, the derivative of $\Theta(\cdot)$ is approximated by a smooth surrogate function $\tilde{\Theta}(\cdot)$ as follows:
\begin{equation}
    \frac{\partial S^t}{\partial V^t} \approx \tilde{\Theta}'(V^t - V_{\mathrm{th}}),
\end{equation}
which allows SNNs to be trained effectively using standard backpropagation through time while preserving their event-driven spiking dynamics.

%% file: sec/03_method.tex
\begin{figure*}[ht]
  \vskip 0.2in
  \begin{center}
    \centerline{\includegraphics[width=0.8\linewidth]{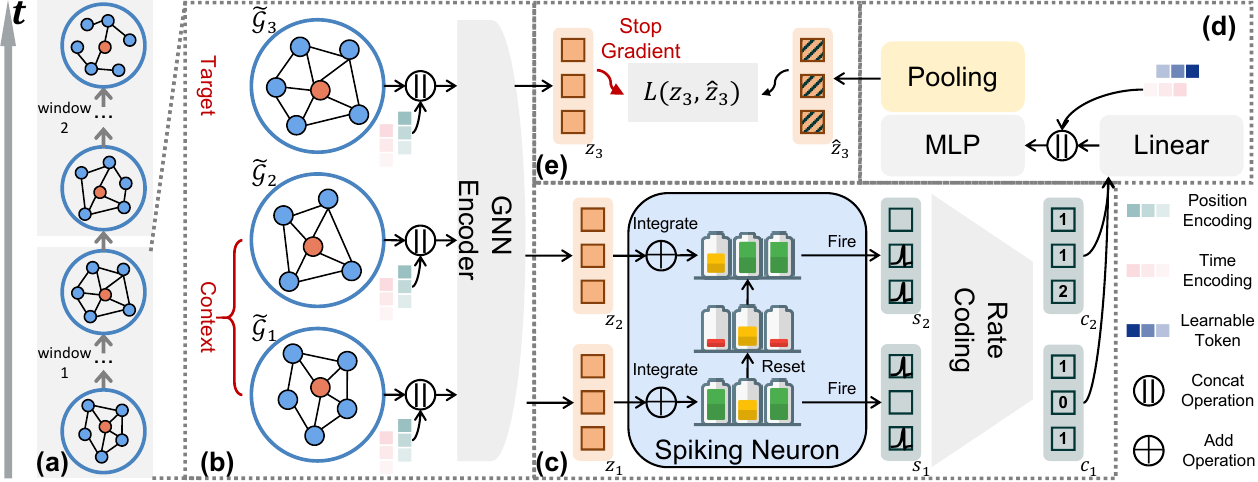}}
    \caption{
      The workflow of the SG-JEPA.
    }
    \label{img:framework}
  \end{center}
\end{figure*}

\section{PRESENT WORK: SG-JEPA}
In this section, we present \textbf{SG-JEPA}, an efficient joint spiking embedding predictive architecture for scalable dynamic graph representation learning. SG-JEPA learns spiking-driven embeddings from context nodes across different graph snapshots and predicts the full-precision representations of target nodes. The overall workflow of SG-JEPA is illustrated in Figure~\ref{img:framework}. It consists of 5 sequential steps: (a) \textbf{Temporal Partitioning} constructs context and target node sets along the temporal dimension; (b) \textbf{Target Encoding} generates full-precision embeddings for target nodes; (c) \textbf{Context Encoding} produces coarse-to-fine spike-count embeddings from context nodes via spiking neural networks; (d) \textbf{Target Representation Prediction}, predicts target embeddings by integrating temporal and spatial contextual information; (e) \textbf{Training Objective}, which optimizes the model through multi-step predictive learning. At last, we describe how SG-JEPA is adapted to downstream tasks.

\subsection{Temporal Partitioning}\label{sec:compnent_tp}
For each node $v \in \mathcal{V}_t$, we sample a fixed number of neighbors at each snapshot to construct a local subgraph. Specifically, let $\mathcal{N}_t(v)$ denote the neighborhood of $v$ at time $t$. We sample a fixed-size set of neighbors for each node to obtain a subgraph $\tilde{\mathcal{G}_t}=\bigcup_{v}^{\mathcal{V}_t}\tilde{\mathcal{N}}_t(v)$, where $\tilde{\mathcal{N}}_t(v)$ is the sampled neighbors. This sampling operator captures the structural evolution along the temporal dimension while maintaining bounded computational cost. 

To enable predictive representation learning, we further partition the snapshot sequence into non-overlapping temporal windows. Given a window size $w$, the snapshot sequence $\tilde{\mathcal{G}}$ is divided into $k=\lfloor T / w \rfloor$ groups:
\begin{equation}
\tilde{\mathcal{G}} \rightarrow \{\tilde{\mathcal{G}}^{(1)}, \tilde{\mathcal{G}}^{(2)}, \ldots, \tilde{\mathcal{G}}^{(k)}\},
\end{equation}
\begin{equation}
\tilde{\mathcal{G}}^{(i)} = \{\tilde{\mathcal{G}}_{(i-1)w+1}, \ldots, \tilde{\mathcal{G}}_{iw}\}, \quad i=1,\ldots,k,
\end{equation}
Within each temporal window $\tilde{\mathcal{G}}^{(i)}$, the nodes appearing in the latest snapshot $\tilde{\mathcal{G}}_{iw}$ are designated as target nodes. The nodes from the preceding snapshots in the same window are treated as context nodes. Formally, for a node $v$, its target representation is defined at time step $iw$, whereas its context information is aggregated from snapshots $\{\tilde{\mathcal{G}}_{(i-1)w+1}, \ldots, \tilde{\mathcal{G}}_{iw-1}\}.$
This temporal partitioning strategy enables SG-JEPA to leverage historical node representations within each window to predict future node representations, while ensuring scalability through fixed-size neighborhood sampling.

\subsection{Target Encoding}\label{sec:tar_encode}
After temporal partitioning, we generate target node representations using graph neural networks, which serve as the predictive targets in the JEPA framework. Relying solely on node embeddings as context or target variables can make the self-predictive objective overly challenging, particularly for dynamic graphs with evolving structures. To alleviate this issue, we incorporate positional and temporal encodings into each sampled subgraph.
The positional encoding helps disambiguate nodes with similar attributes but different topological contexts. The temporal encoding enables the model to distinguish node representations across different snapshots. Moreover, the inner product of temporal encodings can be expressed as the interval between context and target representations. For each target node $v$ at snapshot $t$, we integrate the raw node feature $\mathbf{x} \in \mathbb{R}^{d_x}$, structural positional encoding $\mathbf{p} \in \mathbb{R}^{d_p}$, and temporal encoding $\mathbf{e} \in \mathbb{R}^{d_t}$ into a unified temporal-spatial representation via concatenation:
\begin{equation}
\mathbf{p}_{v,t} = \operatorname{PE}(v \mid \tilde{\mathcal{G}}_t), \quad  \mathbf{e}_t = \operatorname{TE}(t),
\end{equation}
\begin{equation}
\mathbf{h}_{v,t} = [ \mathbf{x}_{v,t} \, \Vert \, \mathbf{p}_{v,t} \, \Vert \, \mathbf{e}_t ],
\end{equation}
where $\Vert$ denotes the concatenation operator. $\operatorname{PE}(\cdot)$ and $\operatorname{TE}(\cdot)$ are positional and temporal encoding functions, respectively. In our implementation, we adopt the random walk positional encoding \cite{dwivedi2021graph} and the Sinusoidal Time Encoding \cite{yu2023towards} to generate additional temporal-spatial information. The integrated representations are then fed into a graph neural network to generate the target node embeddings. Specifically, the node representation $\mathbf{h} \in \mathbb{R}^{d}$ at layer $\ell$ is updated as
\begin{equation}
\mathbf{h}_{v,t}^{\ell} = \operatorname{UPD}^{\ell}\big(\mathbf{h}_{v,t}^{\ell-1}, \operatorname{AGG}^{\ell}(\{\mathbf{h}_{u,t}^{\ell-1} \mid u \in \tilde{\mathcal{N}}_t(v)\})\big),
\end{equation}
where $\operatorname{AGG}(\cdot)$ and $\operatorname{UPD}(\cdot)$ are the aggregation and update operations, respectively. Although we adopt GraphSAGE \cite{hamilton2017inductive} in our implementation, the choice of the graph neural network is not restricted.
The output embedding at the last snapshot of the window, $\mathbf{z}_{v, iw} = \mathbf{h}_{v, iw}^{(L)}$, will be taken as the full-precision target representation of node $v$. This target encoding module produces expressive node embeddings that jointly capture structural context and temporal evolution, providing the prediction objective for the joint embedding predictive framework.

\subsection{Context Encoding}\label{sec:cont_encode}
The context encoder aims to transform sequential full-precision node representations into spiking embeddings that capture temporal dynamics in an efficient and progressive way. In contrast to the target encoder, the context encoder leverages SNNs to generate a sequence of discrete spiking embeddings with increasing temporal resolution. 
For clarity, we just consider the first window here. For the node $v$ within a temporal window, the ordered sequence of full-precision embeddings can be obtained from the target encoder at different snapshots, $\{\mathbf{z}_{v, 1}, \mathbf{z}_{v, 2}, \ldots, \mathbf{z}_{v, w-1}\}$. These embeddings are fed into an SNN to generate spiking counterparts:
\begin{equation}
\mathbf{s}_{v,t} = \mathrm{SNN}(\mathbf{z}_{v,t}), \quad t=1,\ldots,w-1,
\end{equation}
where $\mathbf{s}_{v,t} \in \{0, 1\}^{d}$ represents the spiking embedding. To construct coarse-to-fine contextual representations, the spike count can be obtained via summing $\mathbf{s}_{v,t}$ over multiple time steps:
\begin{equation}
\mathbf{c}_{v,t} = \sum_{t=1}^{w-1} \mathbf{s}_{v,t},
\end{equation}
The spike count sequence $\{\mathbf{c}_{v,1}, \mathbf{c}_{v,2}, \ldots, \mathbf{c}_{v,w-1}\}$ forms a hierarchy of spiking-driven context embeddings. Early elements capture coarse historical information with limited spike accumulation, while later elements progressively integrate more recent spiking activity. The context embeddings closer to the target snapshot incorporate a richer set of spike events and therefore provide finer-grained temporal information. It models temporal dependencies at multiple resolutions, while maintaining the computational efficiency and sparsity benefits of SNNs.

\subsection{Target Representation Prediction}\label{sec:tar_predict}
Given the spike count embeddings from the context encoder, the target representation predictor takes the context representations with multiple precision levels to align the full-precision target representation. Inspired by Matryoshka representation learning \cite{kusupati2022matryoshka}, the nested context embeddings with increasing precision progressively refine the prediction of target node representations. Specifically, we introduce a learnable projection matrix $\mathbf{W} \in \mathbb{R}^{(w-1) d \times d}$ to map the concatenation of the $w$ nested embeddings into a unified latent space.
We define the first $t$ nested embeddings as $\mathbf{c}_{v,1:t}=[\mathbf{c}_{v,1}\Vert \mathbf{c}_{v,2}\Vert \cdots \Vert \mathbf{c}_{v,t}]$. And the projected representation at precision-level $t$ is then computed as
\begin{equation}
\hat{\mathbf{c}}_{v,t} = \mathbf{W}_{1:td}^{\top}\mathbf{c}_{v,1:t}\in\mathbb{R}^{d}.
\end{equation}
where $\mathbf{W}_{1:td}\in\mathbb{R}^{td\times d}$ denotes the sub-matrix formed by the first $td$ rows of $\mathbf{W}$. We concatenate the temporal encoding of the target node to inject  information. Besides, we introduce $w-1$ learnable tokens $\mathbf{m} \in \mathbb{R}^{d}$ which are shared across different time windows to capture global temporal dynamics:
\begin{equation}
\mathbf{u}_{v,t} = [ \hat{\mathbf{c}}_{v, t} \, \Vert \, \mathbf{e}_{w} \, \Vert \, \mathbf{m}_t].
\end{equation}
To obtain a single predictive embedding for each target node, we apply a learnable weighted pooling over predictive representations. The latent target representation at the first window is computed as
\begin{equation}
\hat{\mathbf{z}}_{v,w} = \sum_{t=1}^{w-1} \alpha_t \, \delta (\mathbf{u}_{v, t}),
\end{equation}
where $\alpha$ denotes the learnable pooling weights, and $\delta(\cdot)$ is a shared Multi-Layer Perceptron (MLP) to generate latent target representations.

\subsection{Loss Function}\label{sec:loss}
For each temporal window, SG-JEPA produces a predicted embedding and a corresponding target embedding for every target node.
Here, $\hat{\mathbf{z}}_v^{(i)} \in \mathbb{R}^{d}$ is denoted the predicted embedding of node $v$ in the $i$-th temporal window. And $\mathbf{z}_{v}^{(i)} \in \mathbb{R}^{d}$ be the full-precision target embedding generated from the target encoder. We adopt the InfoNCE objective \cite{gutmann2010noise} to align predicted and target embeddings as follows:
\begin{equation}
\mathcal{D}(\hat{\mathbf{z}}_v^{(i)}, \mathbf{z}_v^{(i)}) = 
- \log
\frac{
\exp\left( \mathrm{sim}\big(\hat{\mathbf{z}}_v^{(i)}, \mathbf{z}_v^{(i)}\big) / \tau \right)
}{
\sum\limits_{u \in \mathcal{V}^{(i)}} 
\exp\left( \mathrm{sim}\big(\hat{\mathbf{z}}_v^{(i)}, \mathbf{z}_u^{(i)}\big) / \tau \right)
},
\end{equation}
where $\mathcal{V}^{(i)}$ denotes the set of target nodes in the $i$-th window. $\tau$ is a temperature hyperparameter and $\mathrm{sim}(\cdot,\cdot)$ is a similarity function. We utilize the stop-gradient operation to prevent the representation collapse following the previous study \cite{chen2020simple}. The overall training objective is accumulated across all windows:
\begin{equation}
\mathcal{L} = \frac{1}{K|\mathcal{V}^{(i)}|} \sum_{i=1}^{K} \sum_{v \in \mathcal{V}^{(i)}} \mathcal{D}\big(\hat{\mathbf{z}}_v^{(i)}, \operatorname{stopgrad}(\mathbf{z}_v^{(i)})\big),
\end{equation}
This objective encourages the predicted embeddings to be close to their corresponding target representations while remaining discriminative across different nodes and temporal windows.

\subsection{Inference Step}\label{sec:inference}
For downstream tasks, SG-JEPA utilizes the last $w-1$ snapshots as the context to predict latent representations without requiring additional training. Concretely, we obtain the sequential full-precision embeddings and transform them into spiking count embeddings $\{\mathbf{c}_{v, (k-1)w+2}, \ldots, \mathbf{c}_{v,kw}\}$. We then obtain the final node representation $\hat{\mathbf{z}}_v$ by learnable pooling as the node representation for downstream tasks. We employ a lightweight MLP as the task-specific classification head.

Notably, SG-JEPA enables flexible precision control for the downstream tasks. At inference time, SG-JEPA does not strictly require all $w-1$ context snapshots. Using fewer context snapshots yields lower-precision yet efficient representations, which are suitable for resource-constrained downstream scenarios.
Since spiking count embeddings and full-precision target embeddings are jointly optimized to predict each other during training, this process implicitly performs self-distillation across temporal resolutions. As a result, lower-precision spike count representations still preserve strong expressiveness.

%% file: sec/04_exp.tex
\section{Experiments}
\paragraph{Datasets} 
To evaluate the effectiveness and scalability of the proposed framework, we conduct experiments on three large-scale temporal graph datasets with different scales: \textbf{DBLP} \cite{lu2019temporal}, \textbf{Tmall} \cite{lu2019temporal} and \textbf{Patent} \cite{file2001lessons}. The datasets
statistics are listed in Table~\ref{tab:dataset_statistic}.

\begin{table}[!ht]
    \centering
    \caption{Dataset Statistics.}\label{tab:dataset_statistic}
    \begin{tabular}{l|cccc}
        \toprule
        \textbf{Datasets} & \textbf{\#Nodes} & \textbf{\#Edges} & \textbf{\#Classes} & \textbf{\#Time Steps} \\
        \midrule
        DBLP   & 28,085    & 236,894    & 10 & 27   \\
        Tmall  & 577,314   & 4,807,545  & 5  & 186  \\
        Patent & 2,738,012 & 13,960,811 & 6  & 25   \\
        \bottomrule
    \end{tabular}
\end{table}

\begin{table*}
    \centering
    \small
    \caption{Macro-F1 scores (\%) on dynamic graphs. The top \first{first}, \second{second} results are highlighted.}\label{tab:macro-f1}
    \begin{tabular}{c|ccc|ccc|ccc}
        \toprule
        \textbf{Methods} & \multicolumn{3}{c|}{\textbf{DBLP}} & \multicolumn{3}{c|}{\textbf{Tmall}} & \multicolumn{3}{c}{\textbf{Patent}} \\
        \midrule
        \textbf{Ratio} & \textbf{40\%} & \textbf{60\%} & \textbf{80\%} & \textbf{40\%} & \textbf{60\%} & \textbf{80\%} & \textbf{40\%} & \textbf{60\%} & \textbf{80\%} \\
        \midrule
        JODIE     & 66.73$_{\pm1.0}$ & 67.32$_{\pm1.0}$ & 67.53$_{\pm1.3}$ & 52.62$_{\pm0.8}$ & 54.02$_{\pm0.6}$ & 54.17$_{\pm0.2}$ & 77.57$_{\pm0.8}$ & 77.69$_{\pm0.6}$ & 77.67$_{\pm0.4}$ \\
        EvolveGCN & 67.22$_{\pm0.3}$ & 69.78$_{\pm0.8}$ & 71.20$_{\pm0.7}$ & 53.02$_{\pm0.7}$ & 54.99$_{\pm0.7}$ & 55.78$_{\pm0.6}$ & 79.67$_{\pm0.4}$ & 79.76$_{\pm0.5}$ &  80.13$_{\pm0.4}$ \\
        TGAT      & 71.18$_{\pm0.4}$ & 71.74$_{\pm0.5}$ & 72.15$_{\pm0.3}$ & 56.90$_{\pm0.6}$ & 57.61$_{\pm0.7}$ & 58.01$_{\pm0.7}$ & 81.51$_{\pm0.4}$ & 81.56$_{\pm0.6}$ & 81.57$_{\pm0.5}$ \\
        ROLAND    & 68.97$_{\pm1.1}$ & 70.53$_{\pm0.3}$ & 71.21$_{\pm1.5}$ & 52.54$_{\pm0.7}$ & 53.67$_{\pm0.3}$ & 53.85$_{\pm0.6}$ & OOM & OOM & OOM \\
        \midrule
        CLDG     & 73.10$_{\pm0.5}$ & 72.83$_{\pm1.2}$ & 73.34$_{\pm0.6}$ & 50.47$_{\pm0.8}$ & 47.31$_{\pm1.2}$ & 44.91$_{\pm0.9}$ & 12.68$_{\pm0.3}$ & 8.07$_{\pm0.8}$ & 6.63$_{\pm0.4}$0.7 \\
        MaskDGNN & 65.40$_{\pm0.7}$ & 66.92$_{\pm0.3}$ & 68.36$_{\pm0.5}$ & 55.90$_{\pm0.2}$ & 55.96$_{\pm0.7}$ & 56.12$_{\pm0.3}$ & OOM & OOM & OOM \\
        \midrule
        GC-SNN     & 70.70$_{\pm0.5}$ & 72.70$_{\pm0.5}$ & 73.07$_{\pm0.3}$ & 58.27$_{\pm0.1}$ & 59.65$_{\pm0.3}$ & 60.43$_{\pm0.3}$ & 82.62$_{\pm0.8}$ & 82.79$_{\pm1.0}$ & 82.92$_{\pm0.9}$ \\
        SpikeNet   & 71.38$_{\pm0.5}$ & 73.28$_{\pm1.0}$ & 74.43$_{\pm0.8}$ & 59.07$_{\pm0.6}$ & 60.84$_{\pm0.5}$ & 62.53$_{\pm0.7}$ & 83.92$_{\pm0.9}$ & 94.03$_{\pm0.9}$ & 94.20$_{\pm0.6}$ \\
        Dy-SIGN    & 70.94$_{\pm0.1}$ & 72.01$_{\pm0.1}$ & 74.67$_{\pm0.5}$ & 57.48$_{\pm0.1}$ & 60.94$_{\pm0.2}$ & 61.89$_{\pm0.1}$ &  83.57$_{\pm0.3}$ & 83.77$_{\pm0.2}$ & 83.91$_{\pm0.2}$ \\
        Delay-DSGN & 72.32$_{\pm0.4}$ & 74.16$_{\pm0.3}$ & 76.54$_{\pm0.4}$ & 60.25$_{\pm0.1}$ & 62.56$_{\pm0.3}$ & 64.02$_{\pm0.2}$ & 83.72$_{\pm0.1}$ & 84.01$_{\pm0.1}$ & 84.20$_{\pm0.1}$ \\
        SiGNN      & \second{73.74$_{\pm0.3}$} & \first{75.96$_{\pm0.2}$} & \second{77.13$_{\pm0.5}$} & \second{61.87$_{\pm0.2}$} & \second{63.90$_{\pm0.3}$} & \first{65.27$_{\pm0.4}$} & \first{84.10$_{\pm0.8}$} & \first{84.38$_{\pm0.3}$} & \first{84.48$_{\pm0.6}$} \\
        \midrule
        SG-JEPA & \first{74.17$_{\pm0.7}$} & \second{75.64$_{\pm0.4}$} & \first{77.21$_{\pm0.3}$} & \first{61.89$_{\pm0.2}$} & \first{63.94$_{\pm0.3}$} & \second{64.85$_{\pm0.2}$} & \second{83.84$_{\pm0.1}$} & \second{84.33$_{\pm0.1}$} & \second{84.37$_{\pm0.3}$} \\
        \bottomrule
    \end{tabular}
\end{table*}

\begin{table*}
    \centering
    \small
    \caption{Micro-F1 scores (\%) on dynamic graphs. The top \first{first}, \second{second} results are highlighted.}\label{tab:micro-f1}
    \begin{tabular}{c|ccc|ccc|ccc}
        \toprule
        \textbf{Methods} & \multicolumn{3}{c|}{\textbf{DBLP}} & \multicolumn{3}{c|}{\textbf{Tmall}} & \multicolumn{3}{c}{\textbf{Patent}} \\
        \midrule
        \textbf{Ratio} & \textbf{40\%} & \textbf{60\%} & \textbf{80\%} & \textbf{40\%} & \textbf{60\%} & \textbf{80\%} & \textbf{40\%} & \textbf{60\%} & \textbf{80\%} \\
        \midrule
        JODIE     & 68.44$_{\pm0.6}$ & 68.51$_{\pm0.8}$ & 68.80$_{\pm0.9}$ & 58.36$_{\pm0.5}$ & 60.28$_{\pm0.3}$ & 60.49$_{\pm0.3}$ & 77.64$_{\pm0.7}$ & 77.89$_{\pm0.5}$ & 77.97$_{\pm0.4}$ \\
        EvolveGCN & 69.12$_{\pm0.8}$ & 70.43$_{\pm0.6}$ & 71.32$_{\pm0.5}$ & 59.96$_{\pm0.7}$ & 61.19$_{\pm0.6}$ & 61.77$_{\pm0.6}$ & 79.39$_{\pm0.5}$ & 79.75$_{\pm0.3}$ & 80.01$_{\pm0.3}$ \\
        TGAT      & 71.10$_{\pm0.2}$ & 71.85$_{\pm0.4}$ & 73.12$_{\pm0.3}$ & 62.05$_{\pm0.5}$ & 62.92$_{\pm0.4}$ & 93.32$_{\pm0.7}$ & 80.79$_{\pm0.7}$ & 80.81$_{\pm0.6}$ & 80.93$_{\pm0.6}$ \\
        ROLAND    & 70.16$_{\pm0.7}$ & 71.11$_{\pm0.4}$ & 71.95$_{\pm1.8}$ & 59.91$_{\pm0.5}$ & 61.08$_{\pm0.1}$ & 61.24$_{\pm0.3}$ & OOM & OOM & OOM \\
        \midrule
        CLDG     & \first{75.51$_{\pm0.3}$} & 75.01$_{\pm0.6}$ & 75.49$_{\pm0.4}$ & 53.77$_{\pm1.0}$ & 52.03$_{\pm0.6}$ & 51.32$_{\pm1.0}$ & 23.49$_{\pm0.1}$ & 23.27$_{\pm0.5}$ & 23.25$_{\pm0.5}$ \\
        MaskDGNN & 66.30$_{\pm0.5}$ & 67.10$_{\pm0.4}$ & 68.86$_{\pm0.6}$ & 58.81$_{\pm0.2}$ & 59.17$_{\pm0.4}$ & 59.19$_{\pm0.5}$ & OOM & OOM & OOM \\
        \midrule
        GC-SNN     & 71.73$_{\pm0.5}$ & 73.86$_{\pm0.3}$ & 74.26$_{\pm0.8}$ & 62.60$_{\pm0.3}$ & 64.14$_{\pm0.2}$ & 64.65$_{\pm0.4}$ & 82.52$_{\pm0.8}$ & 82.65$_{\pm0.7}$ & 82.84$_{\pm0.6}$ \\
        SpikeNet   & 72.03$_{\pm0.3}$ & 74.18$_{\pm0.6}$ & 75.72$_{\pm0.3}$ & 63.43$_{\pm0.6}$ & 65.13$_{\pm1.0}$ & 66.51$_{\pm0.5}$ & 83.89$_{\pm0.7}$ & 83.99$_{\pm0.8}$ & 84.18$_{\pm0.2}$ \\
        Dy-SIGN    & 71.90$_{\pm0.1}$ & 72.61$_{\pm0.4}$ & 74.96$_{\pm0.2}$ & 62.93$_{\pm0.3}$ & 64.10$_{\pm0.3}$ & 65.82$_{\pm0.2}$ & 83.50$_{\pm0.2}$ & 83.47$_{\pm0.1}$ & 83.90$_{\pm0.2}$ \\
        Delay-DSGN & 72.56$_{\pm0.2}$ & 74.44$_{\pm0.3}$ & 76.87$_{\pm0.5}$ & 64.32$_{\pm0.1}$ & 66.20$_{\pm0.2}$ & 67.88$_{\pm0.4}$ & 83.66$_{\pm0.1}$ & 83.97$_{\pm0.1}$ & 84.15$_{\pm0.1}$ \\
        SiGNN      & 74.66$_{\pm0.7}$ & \second{76.46$_{\pm0.2}$} & \second{77.98$_{\pm1.0}$} & \first{65.74$_{\pm0.2}$} & \first{67.38$_{\pm0.1}$} & \first{68.58$_{\pm0.2}$} & \first{84.06$_{\pm0.5}$} & \first{84.34$_{\pm0.9}$} & \first{84.45$_{\pm0.2}$} \\
        \midrule
        SG-JEPA & \second{74.86$_{\pm0.4}$} & \first{76.68$_{\pm0.4}$} & \first{78.00$_{\pm0.1}$} & \second{65.36$_{\pm0.2}$} & \second{67.22$_{\pm0.2}$} & \second{68.24$_{\pm0.1}$} & \second{83.80$_{\pm0.2}$} & \second{84.26$_{\pm0.4}$} & \second{84.30$_{\pm0.7}$} \\
        \bottomrule
    \end{tabular}
\end{table*}

\paragraph{Baselines} 
We compare SG-JEPA with a diverse set of representative dynamic graph models spanning full-precision, self-supervised and spiking paradigms. Specifically, we include four full-precision supervised or semi-supervised DGNNs, namely JODIE \cite{kumar2019predicting}, EvolveGCN \cite{pareja2020evolvegcn}, TGAT \cite{xu2020inductive} and ROLAND \cite{you2022roland}. We introduce two representative self-supervised baselines, CLDG \cite{xu2023cldg} and MaskDGNN \cite{he2025maskdgnn}. There are five spiking DGNN baselines, including GC-SNN \cite{xu2021exploiting}, SpikeNet \cite{li2023scaling}, Dy-SIGN \cite{yin2024dynamic}, Delay-DSGN \cite{wang2025delay} and SiGNN \cite{chen2025signn}. 
For all baselines, we conduct a hyperparameter search to obtain optimal configurations on each dataset. Each experiment is repeated five times with different random seeds. We report the average \textbf{Macro-F1} and \textbf{Micro-F1} scores. We further assess performance under three different training ratios following the previous study \cite{li2023scaling}.

\subsection{Overall Performance}
Table~\ref{tab:macro-f1} and Table~\ref{tab:micro-f1} report the Macro-F1 and Micro-F1 results of all baselines on DBLP, Tmall and Patent under different training ratios. Overall, \textbf{SG-JEPA achieves performance comparable to state-of-the-art supervised spiking DGNNs and consistently outperforms existing self-supervised baselines in most settings}.  Across all datasets and training ratios, SG-JEPA delivers competitive and stable performance, demonstrating the effectiveness of joint-embedding predictive learning for scalable dynamic graph representation learning with SNNs.

Compared with full-precision supervised DGNNs, spiking-based models achieve comparable or even superior performance, particularly on datasets with pronounced temporal dynamics such as DBLP and Tmall. This observation suggests that spiking neural networks are effective in modeling temporal evolution in dynamic graphs through event-driven computation and temporal accumulation. Despite operating with discrete spike-based representations, \textbf{SG-JEPA achieves average relative improvements of 6.6\% and 5.6\% in Macro-F1 and Micro-F1, respectively, over full-precision baselines across all datasets.} Moreover, The average performance gap between SG-JEPA and the state-of-the-art supervised spiking baselines, SiGNN, is only \textbf{0.085\%}. It indicates nearly identical performance.
SG-JEPA further demonstrates clear advantages in both effectiveness and scalability when compared with existing contrastive-based and generative-based dynamic graph methods. CLDG relies on contrastive objectives that are sensitive to training stability and data scale. Although we strictly follow the official implementation and employ an MLP-based classification head, CLDG exhibits unstable optimization and degraded performance on large-scale dynamic graphs. In contrast, SG-JEPA avoids explicit negative sampling and manually designed graph augmentations, resulting in more stable training and consistent performance gains. Moreover, MaskDGNN adopts an edge-level reconstruction objective that incurs substantial memory overhead and becomes impractical on large graphs, leading to out-of-memory (OOM) issues on the Patent dataset. By contrast, SG-JEPA scales gracefully while maintaining competitive performance. Finally, compared with existing spiking DGNNs, \textbf{SG-JEPA achieves consistent improvements despite being trained in a self-supervised manner, yielding average improvement of 1.6\% and 1.4\% in Macro-F1 and Micro-F1 across all datasets, respectively.} These results indicate that JEPA effectively enhances spiking-driven representation, enabling strong performance without reliance on labeled supervision.

\subsection{Training Efficiency Analysis}
As we mentioned above, though existing self-supervised methods may achieve promising performance, they often incur substantial computational and memory overhead and may exhibit misalignment between pretext objectives and downstream node-level tasks. As a result, improvements in self-supervised training do not always translate into consistent gains in downstream performance. SG-JEPA mitigates these issues via a joint-embedding predictive objective that operates directly in embedding space and captures temporal dynamics with a lightweight spiking neural network. Here, we investigate two key questions to compare SG-JEPA with existing self-supervised dynamic graph methods based on several efficiency metrics.

\emph{\textbf{Q1:} Does eliminating augmentation and reconstruction significantly improve training efficiency?}

\begin{figure}[!ht]
  \begin{center}
    \centering
    \includegraphics[width=\linewidth]{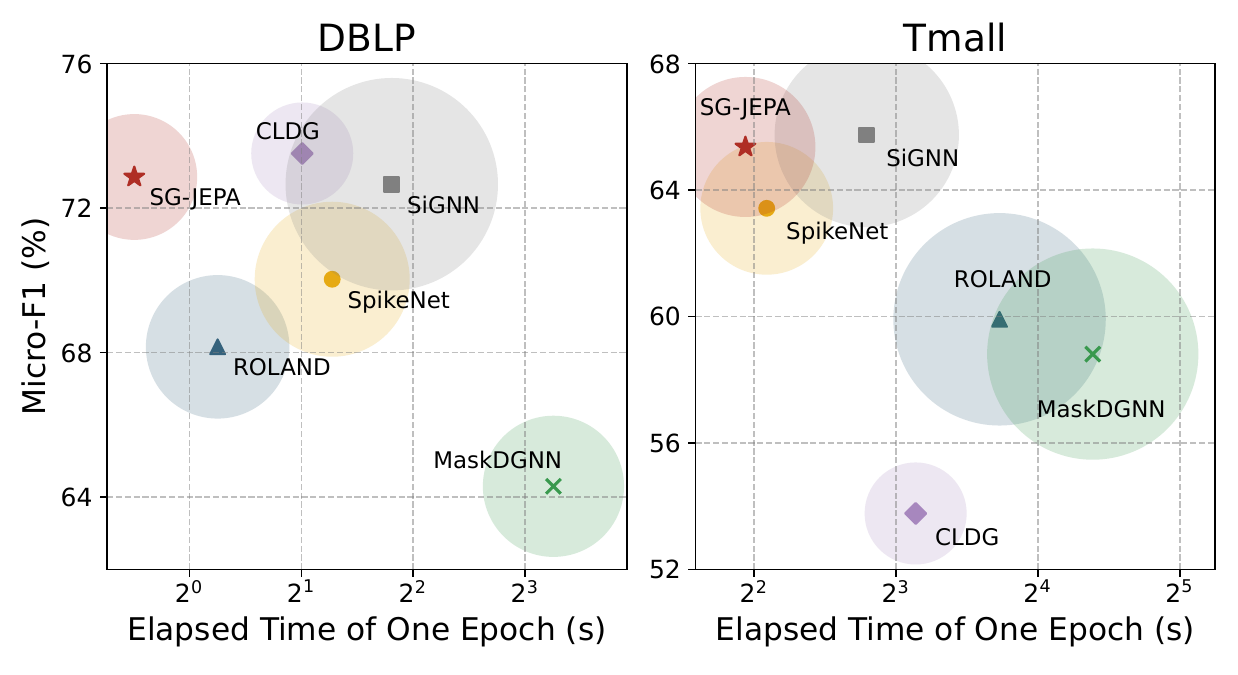}
    \caption{Comparison of per-epoch training time and Micro-F1 across different methods. The size of the circle indicates the maximum memory usage during the training.}
    \label{fig:training_efficiency}
  \end{center}
  % \vspace{-10px}
\end{figure}

We measure two efficiency-related metrics during training: (i) the elapsed time of one training epoch and (ii) the maximum memory usage. Figure~\ref{fig:training_efficiency} demonstrates the comparison of Micro-F1 and per-epoch training time across different baselines. 
Across both DBLP and Tmall datasets, SG-JEPA achieves substantially faster training. Specifically, compared to the average elapsed time across other baselines, \textbf{SG-JEPA achieves a 5.2$\times$ speedup on DBLP and a 2.8$\times$ speedup on Tmall in terms of per-epoch training time, while maintaining competitive Micro-F1 performance.} This efficiency gain can be attributed to the simplified training pipeline of SG-JEPA, which performs joint-embedding prediction directly in latent space and avoids explicit edge-level reconstruction and multi-view graph augmentation. By contrast, contrastive and reconstruction-based methods incur substantial computational overhead due to complex machinery (e.g., augmentation and negative sampling). Regarding memory usage, SG-JEPA exhibits competitive efficiency, ranking second only to CLDG. In contrast, reconstruction-based methods such as MaskDGNN demonstrate significantly larger memory footprints due to edge-level decoding. These results confirm that the predictive learning paradigm combined with spike dynamics substantially reduces both training time and memory overhead without sacrificing performance.

\begin{figure}[!ht]
  \begin{center}
    \centering
    \includegraphics[width=\linewidth]{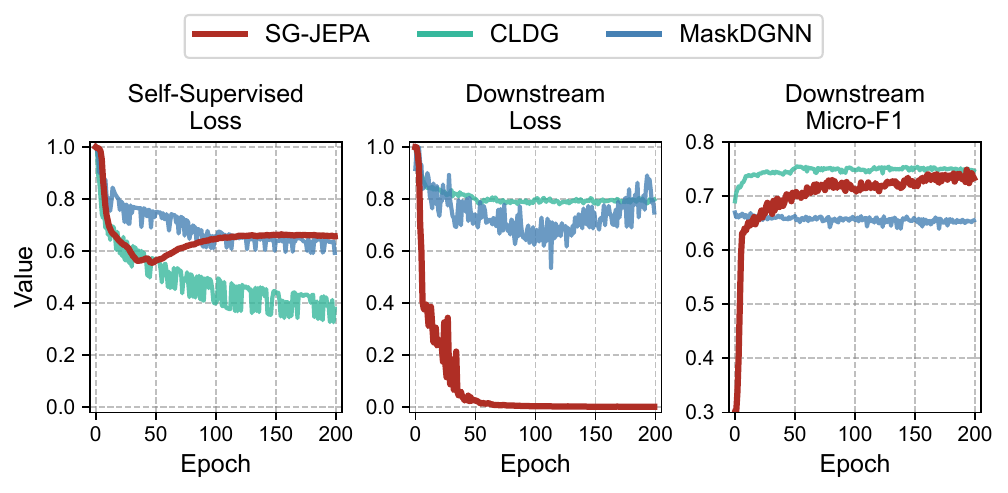}
    \caption{ Learning curve of three self-supervised methods.}
    \label{fig:convergence_analysis}
  \end{center}
  % \vspace{-10px}
\end{figure}

\emph{\textbf{Q2:} Does SG-JEPA avoid downstream task misalignment and representation collapse?}

To analyze convergence behavior, we normalize self-supervised loss and downstream node classification loss to the same scale and monitor their evolution alongside downstream Micro-F1. Figure~\ref{fig:convergence_analysis} illustrates the relationship among pretext loss, downstream loss and downstream performance across training epochs. As shown in Figure~\ref{fig:convergence_analysis}, the self-supervised predictive loss decreases smoothly while maintaining stable downstream improvements. It indicates that SG-JEPA learning dynamic representations stably without signs of collapse. In contrast, MaskDGNN optimizes an edge-level reconstruction objective whose loss continues to decrease while downstream performance degrades, highlighting a potential misalignment between the pretext objective and node-level tasks. We notice that SG-JEPA converges in a similar number of epochs as CLDG without well-designed multi-view augmentations. The downstream performance of SG-JEPA closely matches that of CLDG, while employing a substantially simpler training pipeline. These results indicate that JEPA can achieve stable convergence and competitive downstream performance without contrastive view construction, while avoiding the degradation observed in reconstruction-based methods.

\begin{figure}[!ht]
  \begin{center}
    \centering
    \includegraphics[width=\linewidth]{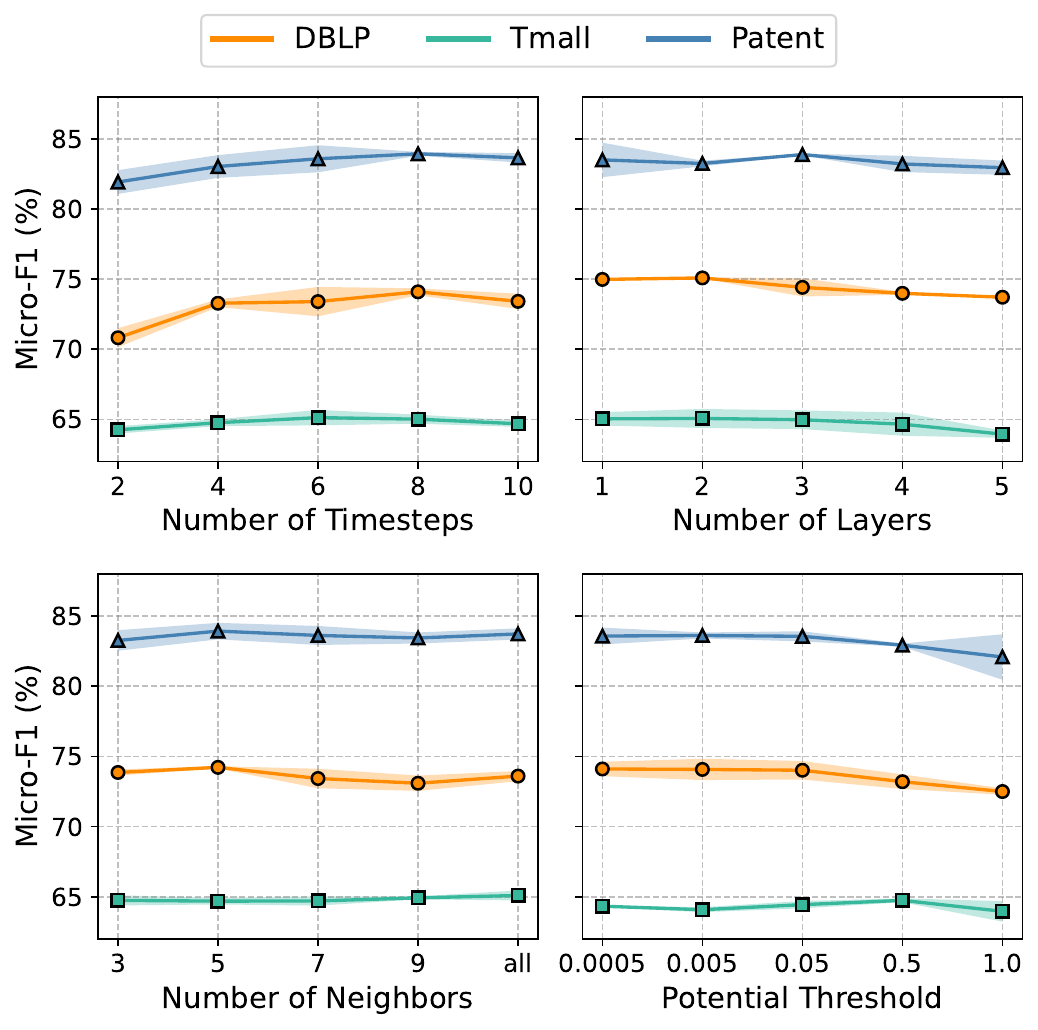}
    \caption{Parameter sensitivity analysis on the DBLP, Tmall and Patent datasets.}
    \label{fig:sensitivity}
  \end{center}
\end{figure}

\subsection{Parameter Sensitivity Analysis}
We evaluate the robustness of SG-JEPA with respect to four key hyperparameters: window size, the number of GNN layers, the number of sampled neighbors and the membrane potential threshold of the spiking neuron. For each hyperparameter, we vary one factor while keeping the others fixed at their default settings. Experiments are conducted on DBLP, Tmall and Patent, and we utilize Micro-F1 as the evaluation metric. The experimental results are illustrated in Figure~\ref{fig:sensitivity}.

As depicted in the Figure~\ref{fig:sensitivity}, SG-JEPA demonstrates stable performance across a wide range of configurations. The window size determines the temporal resolution of spike-count embeddings.  When the window size is very small, $w=2$, the model will degenerate into binary spike encoding during the training. For this extreme case,  quantization loss results in noticeable performance degradation. Notably, benefiting from the nested spike-count embedding design, SG-JEPA achieves competitive performance with only a few time steps, matching or surpassing spiking baselines without requiring long temporal contexts. Increasing the number of layers or the number of sampled neighbors does not monotonically improve results. Deeper architectures are susceptible to over-smoothing on large-scale dynamic graphs, while sampling all neighbors may incur the over-squashing \cite{alon2020bottleneck}. The membrane potential threshold regulates spike sparsity by controlling neuron firing frequency. Excessively high thresholds suppress spiking activity and hinder information propagation. Overall, the parameter sensitivity analysis shows that SG-JEPA delivers strong and consistent performance under diverse configurations, while preserving a promising balance between representational expressiveness and computational efficiency.

\subsection{Ablation Study}
To uncover the contribution of each design choice in SG-JEPA, ablation studies are conducted on the DBLP and Tmall datasets. The results are summarized in Table~\ref{tab:ablation}. Specifically, we consider three different ablation studies: (1) \emph{w/} IF and \emph{w/} LIF, where the proposed PLIF neuron is replaced by IF or LIF neurons; (2) \emph{w/} EMA and \emph{w/o} stopgrad, which aim to investigate different target encoder update strategies. For the former, it follows standard JEPA practice by removing the stop-gradient operation and updating a target encoder by exponential moving average. (3) \emph{w/} $\ell_{1}$, \emph{w/} $\ell_{2}$ and \emph{w/} Cosine, where different loss functions are adopted to measure the discrepancy between the predicted and target embeddings.

We first examine the influence of different spiking neurons. As shown in Table~\ref{tab:ablation}, replacing the proposed PLIF neuron with IF or LIF consistently degrades performance. It highlights the importance of learnable membrane dynamics for capturing temporal dependencies. For the choice of prediction objective, common distance losses result in inferior performance compared with the default formulation, suggesting that preserving directional consistency in embedding space is beneficial for predictive latent representations.
Besides, a natural intuition is that the InfoNCE objective provides implicit anti-collapse regularization, which may allow the target encoder to be unfrozen during training. However, the results in Table~\ref{tab:ablation} show that removing the stop-gradient constraint leads to a substantial performance drop. We believe that asymmetric gradient flow remains critical for stable training in spiking JEPA-style models. 
Introducing EMA may partially mitigate this degradation and yield more stable performance.

\begin{table}[thb]\small
    \centering
    \caption{Ablation results (\%) on DBLP and Tmall datasets.}
    \begin{tabular}{l|cc|cc|c}
        \toprule
        \multirow{2}{*}{\textbf{Models}} & \multicolumn{2}{c|}{\textbf{DBLP}} & \multicolumn{2}{c|}{\textbf{Tmall}} & \multirow{2}{*}{\textbf{$\Delta$}} \\
        \cmidrule(lr){2-3} \cmidrule(lr){4-5}
        & Macro-F1 & Micro-F1 & Macro-F1 & Micro-F1 & \\
        \midrule
        SG-JEPA & \second{74.17$_{\pm0.7}$} & \first{74.86$_{\pm0.4}$} & \first{61.89$_{\pm0.2}$} & \first{65.36$_{\pm0.2}$} & -- \\
        \midrule
        \emph{w/} IF  & 71.25$_{\pm0.2}$ & 71.98$_{\pm0.2}$ & \second{61.23$_{\pm0.1}$} & \second{64.35$_{\pm0.1}$} & -1.9 \\
        \emph{w/} LIF & 73.11$_{\pm0.3}$ & 73.54$_{\pm0.2}$ & 59.78$_{\pm0.7}$ & 63.84$_{\pm0.1}$ & -1.5\\
        \midrule
        \emph{w/o} stopgrad & 71.73$_{\pm0.2}$ & 72.23$_{\pm0.9}$ & 58.44$_{\pm0.5}$ & 63.68$_{\pm0.3}$ & -2.5 \\
        \emph{w/} EMA & 73.11$_{\pm0.1}$ & 73.72$_{\pm0.2}$ & 60.61$_{\pm0.4}$ & 64.03$_{\pm0.3}$ & -1.2 \\
        \midrule
        \emph{w/} L1     & \first{74.26$_{\pm0.1}$} & \second{74.71$_{\pm0.1}$} & 58.60$_{\pm0.2}$ & 62.97$_{\pm0.1}$ & -1.4\\
        \emph{w/} L2     & 73.23$_{\pm0.2}$ & 73.81$_{\pm0.1}$ & 60.52$_{\pm0.2}$ & 64.28$_{\pm0.3}$ & -1.1\\
        \emph{w/} Cosine & 70.45$_{\pm0.0}$ & 71.69$_{\pm0.2}$ & 56.65$_{\pm0.5}$ & 61.72$_{\pm0.3}$ & -3.9\\
        \bottomrule
    \end{tabular}
    \label{tab:ablation}
\end{table}

%% file: sec/05_concl.tex
\section{Conclusion}
In this work, we introduce \textbf{SG-JEPA}, the first attempt to bring the joint-embedding predictive paradigm to dynamic graph representation learning. For graph data, Conventional self-supervised learning frameworks typically rely heavily on augmentation and reconstruction processes, which often become computational bottlenecks in real-world deployments. SG-JEPA performs predictive learning directly in embedding space, substantially reducing computational overhead. Our results indicate that latent-space prediction offers a promising and scalable alternative for self-supervised learning on large dynamic graphs and merits further exploration.